\documentclass[a4paper,fleqn]{cas-sc}

\usepackage[numbers]{natbib}

\def\tsc#1{\csdef{#1}{\textsc{\lowercase{#1}}\xspace}}
\tsc{WGM}
\tsc{QE}
\tsc{EP}
\tsc{PMS}
\tsc{BEC}
\tsc{DE}

\usepackage{times}
\usepackage{epsfig}
\usepackage{graphicx}
\usepackage{amsmath}
\usepackage{amssymb}
\usepackage{footnote}
\usepackage{dsfont}
\usepackage{enumitem}

\usepackage{comment}
\usepackage{blindtext}
\usepackage{morewrites}
\usepackage{makecell}
\usepackage{pifont}%
\newcommand{\xmark}{\ding{55}}%
\newcommand{\cmark}{\ding{51}}%
\usepackage{multirow}
\usepackage{xcolor}
\usepackage{float}

\usepackage[utf8]{inputenc} %
\usepackage[T1]{fontenc}    %
\usepackage{hyperref}       %
\usepackage{url}            %
\usepackage{booktabs}       %
\usepackage{amsfonts}       %
\usepackage{nicefrac}       %
\usepackage{microtype}      %
\usepackage{xcolor}         %

\usepackage[utf8]{inputenc} %
\usepackage[T1]{fontenc}    %
\usepackage{hyperref}       %
\usepackage{url}            %
\usepackage{booktabs}       %
\usepackage{amsfonts}       %
\usepackage{nicefrac}       %
\usepackage{microtype}      %
\usepackage{xcolor}         %

\usepackage{wrapfig}
\usepackage{graphicx}
\usepackage{amsmath}
\usepackage{mathtools}
\usepackage{multirow}
\usepackage{makecell}
\usepackage{tabularx}
\usepackage{algorithm2e}
\usepackage{fancyvrb,xcolor}
\usepackage{nicematrix}
\usepackage{bbm}

\usepackage{algorithm2e}
\usepackage{fancyvrb}
\usepackage{xcolor}
\usepackage{nicematrix}

\usepackage{tikz}
\usepackage{comment}
\usepackage{amsmath,amssymb} %
\usepackage{color}
\usepackage{enumitem}
\usepackage{amsthm}

\usepackage{multirow}
\usepackage{makecell}
\usepackage{amsmath}
\usepackage{capt-of}
\usepackage{tabularx}
\usepackage{epsfig}
\usepackage{amssymb}
\usepackage{amsfonts}
\usepackage{booktabs}
\usepackage{scalerel}
\usepackage{listings}
\usepackage{varwidth}
\usepackage[export]{adjustbox}
\usepackage{tikz}
\usetikzlibrary{tikzmark}

\usepackage{stmaryrd}
\usepackage{bbm}
\usepackage{wrapfig}
\usepackage{pifont}
\usepackage[utf8]{inputenc}

\usepackage{graphicx}%
\usepackage{multirow}%
\usepackage{makecell}
\usepackage{tabularx}
\usepackage{multicol}
\usepackage{tabulary}
\usepackage{amsmath,amssymb,amsfonts}%
\usepackage{amsthm}%
\usepackage{mathrsfs}%
\usepackage{xcolor}
\usepackage{textcomp}%
\usepackage{manyfoot}%
\usepackage{booktabs}%
\usepackage{algpseudocode}%
\usepackage{listings}%
\usepackage{caption}
\usepackage{booktabs}
\usepackage{pifont}
\usepackage{footnote}

\usepackage{xspace}

\usepackage{tcolorbox}
\definecolor{mygreen}{HTML}{3cb44b}
\newcommand{\Bac}[1]{\textcolor{black}{#1}}

\DontPrintSemicolon

\SetKwComment{Comment}{\color{green!50!black}\# }{}

\SetKwProg{Function}{def}{:}{}

\SetKwProg{For}{for}{:}{}
\SetKwProg{If}{if}{:}{}

\begin{document}
\let\WriteBookmarks\relax
\def\floatpagepagefraction{1}
\def\textpagefraction{.001}
\shorttitle{BRAIN: Bias-Mitigation Continual Learning Approach to Vision-Brain Understanding}
\shortauthors{X.B Nguyen et~al.}

\title [mode = title]{BRAIN: Bias-Mitigation Continual Learning Approach to Vision-Brain Understanding}

\author[1]{Xuan-Bac Nguyen}
\author[1]{Thanh-Dat Truong}%
\author[2]{Pawan Sinha}
\author[1]{Khoa Luu}

\affiliation[1]{organization={
Electrical Engineering \& Computer Science Department, University of Arkansas},
                addressline={1 University of Arkansas}, 
                postcode={72703 AR}, 
                postcodesep={}, 
                city={Fayetteville},
                country={USA}}

\affiliation[2]{organization={
Department of Brain and Cognitive Sciences, Massachusetts Institute of Technology},
                addressline={77 Massachusetts Ave}, 
                postcode={02139 MA}, 
                postcodesep={}, 
                city={Cambridge},
                country={USA}}

\begin{abstract}
Memory decay makes it harder for the human brain to recognize visual objects and retain details. Consequently, recorded brain signals become weaker and more uncertain, and they contain poorer visual context over time. This paper presents one of the first vision-learning approaches to address this problem. First, we statistically and experimentally demonstrate the existence of inconsistency in brain signals and its impact on the Vision-Brain Understanding (VBU) model. Our findings show that brain signal representations shift across recording sessions, leading to compounding bias that poses challenges for model learning and degrades performance. Then, we propose a new Bias-Mitigation Continual Learning (BRAIN) approach to address these limitations. In this approach, the model is trained \Bac{in a continual learning setup, mitigating the growing bias from each learning step}. A new loss function, De-bias Contrastive Learning, is also introduced to address the bias problem. In addition, to prevent catastrophic forgetting, where the model loses knowledge from previous sessions, the new Angular-based Forgetting Mitigation approach is introduced to preserve learned knowledge. Finally, the empirical experiments demonstrate that our approach achieves State-of-the-Art (SOTA) performance across various benchmarks, surpassing prior and non-continual learning methods.
\end{abstract}


\maketitle

\section{Introduction}

\begin{figure*}[!ht]
    \centering
    \includegraphics[width=1.0\linewidth]{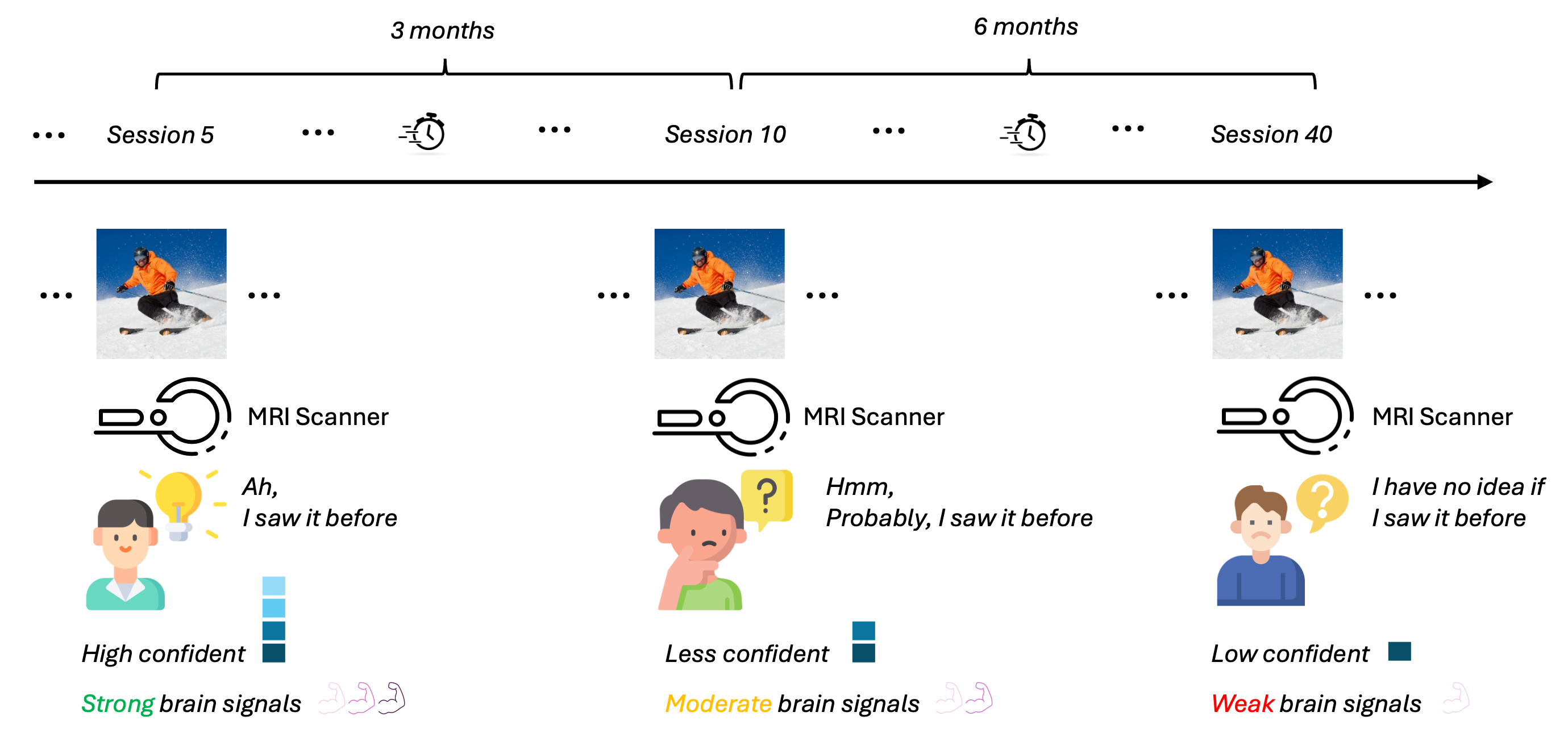}
    \caption{Participant confidence in recognizing visual stimuli declined over time. Consequently, brain signals from earlier sessions were stronger and contained richer visual context than later sessions. Its shift introduces a change in distribution and bias in visual brain signals. While previous studies have overlooked this problem, this work is the first to highlight it.}
    \label{fig:problem_introduction}
\end{figure*}

The human brain is a complex system, with about one-third of its surface dedicated to processing visual information, enabling interpretation, object recognition, and scene understanding \cite{60_van1992information, 59_tsao2006cortical, 38_liang2018fine, 62_wang1996optical}. Researchers study this process using brain-scanning techniques such as fMRI to gain insights into the relationship between vision and brain function \cite{50_raichle2001default, 27_hubel1959receptive, 45_nauhaus2012orthogonal, 40_livingstone1984anatomy, 46_posner1980attention, 26_hubel1968receptive}. Recently, large-scale datasets have been introduced to advance research on vision and brain function \cite{nsd, things, neuromod}. Unlike simple collections of instance records, these datasets are built over extended periods, often months or years. For instance, the Natural Scenes Dataset (NSD) took twelve months to complete. During data collection, participants may be exposed to visual stimuli they encountered weeks or months earlier.  Prior studies \cite{6_block2011perceptual, 7_buschman2011neural, 13_cohen2016bandwidth, 47_pylyshyn1999vision, 68_whitney2018ensemble} on human perception and cognition have shown that the amount of visual information a person can process and retain at any moment is limited by the capacity of visual working memory \cite{43_luck2013visual} and varies between individuals. 
Due to the natural process of \textit{memory decay}, individuals may struggle to recall previously seen stimuli, leading to lower confidence in their responses as demonstrated in Fig. \ref{fig:problem_introduction}. Consequently, fMRI signals recorded in later sessions may exhibit decreased visual contextual information compared to those from earlier sessions.

\noindent
\textbf{Limitations in Prior Work.} Vision-brain understanding (VBU) has recently made significant strides in decoding brain signals \cite{ozcelik2022reconstruction, shen2019deep, takagi, mindeye1, mindbridge, psychometry, neuropictor, mindeye2}. These methods extract visual context hidden within brain signals, often by retrieving or reconstructing the original visual stimuli using generative GANs or Diffusion models. However, they typically treat all data points in the dataset as equal, overlooking the importance of bias in brain responses \cite{mindeye1, mindeye2, xia-umbrae-2024}. At the same time, they combine all data sessions to train. This limitation can make it challenging for models to learn from fMRI signals recorded in later sessions, where these signals may reflect a less certain visual context and be less consistent with earlier data. This problem may lead to a performance degradation in vision-brain models (details will be provided in Section \ref{sec:fmri_shift_drift}). This paper aims to address the non-stationarity of brain signals arising from human \textit{memory decay}. 

\noindent
\textbf{Contributions of this Work.} Our contributions are summarized as follows\footnote{The source code of this work will be publicly available.}.
\begin{itemize}
    \item First, we statistically and experimentally uncover the bias problem in brain signals and its impact on the performance of Vision-Brain models, a problem that most prior studies have overlooked.  
    \textit{To the best of our knowledge, we present one of the first studies to address this bias problem}.
    
    \item \Bac{Second, in a continual VBU setup, brain signals become increasingly inconsistent over time, i.e., across data collection sessions, leading to representation bias in later sessions or training steps. To address this problem, we propose a novel Bias-Mitigation Continual Learning (BRAIN) approach that incrementally learns brain signals throughout the data collection. This approach is designed to mitigate bias and reduce variance in the data}.
    
    \item Third, we introduce a new \textit{De-bias Contrastive Learning (DCL) loss} to mitigate signal bias and a new \textit{Angular-based Forgetting Mitigation (AFM) loss} to prevent catastrophic forgetting.
    
    \item Finally, the empirical results demonstrate the performance improvements of the proposed approach, surpassing previous methods across various continual learning benchmarks.
\end{itemize}

\section{Related Work}

\subsection{Vision-Brain Understanding}
Neural decoding involves interpreting neural signals, e.g., EEG, MEG, fMRI, to infer human perception and cognitive states. Recent advancements have led to substantial progress in this domain, particularly in applications like motor imagery decoding \cite{2_aflalo2015decoding} and visual decoding \cite{3_bai2023dreamdiffusion, nguyen2023micron, nguyen2021clusformer, nguyen2020self,nguyen2022multi,nguyen2024insect,nguyen2024qclusformer,truong2022otadapt,nguyen2024diffusion,nguyen2023quantum, nguyen2022two, nguyen2024quantumrevisied, nguyen2023fairness, nguyen2023brainformer, nguyen2019sketch, xia2025exploring, nguyen2023algonauts,nguyen2019audio, nguyen2024hierarchical,truong2025insect,nguyen2024quantum, nguyen2024bractive, nguyen2024cobra, 4_beliy2019voxels, 18_fang2023alleviating, 2_aflalo2015decoding}. Visual decoding consists of two main tasks: brain-to-image retrieval and reconstruction. Various methods \cite{10_chen2024visual, 15_du2023decoding, 36_li2024visual, 56_song2023decoding} have been introduced to align EEG/MEG signal representations with the Contrastive Vision-Language Pre-training (CLIP) \cite{49_radford2021learning} embedding space. However, they fail to address the inherent mismatch between brain signals and visual stimuli, leading to overfitting on the training set and poor generalization to new data. In Vision-Brain Understanding (VBU), fMRI signals recorded while subjects view an image serve as input for reconstructing the initially observed image \cite{neuropictor, mindeye1, mindeye2, psychometry, mindbridge, takagi, xia-umbrae-2024}. 
Several studies have utilized diffusion models \cite{thermodynamics_2015, ho-denoising-2020, openai-diffusion-2021, rombach-highresolutionimagesynthesislatent-2022, xu-versatilediffusiontextimages-2024} to model fMRI signal distributions. 
MindEye \cite{mindeye1} employs contrastive learning to align image and fMRI latent spaces before using diffusion models for image reconstruction. 
MindEye2 \cite{mindeye2} adapts to a limited data setting to mitigate the high cost of fMRI data collection. 
Takagi et al. \cite{takagi} apply latent diffusion \cite{rombach-highresolutionimagesynthesislatent-2022} for image reconstruction, conditioning on specific fMRI segments. 
MindBridge \cite{mindbridge} introduces a cyclic fMRI reconstruction strategy to align brain data across subjects, facilitating robust brain-to-image and text decoding via dual embeddings. 
Psychometry \cite{psychometry}, in contrast, employs a unified model to extract common and individual features across all subjects. 
NeuroPictor \cite{neuropictor} integrates a latent diffusion-like mechanism \cite{rombach-highresolutionimagesynthesislatent-2022} to encode fMRI signals into a latent space, distinguishing between high- and low-level features for improved fine-detail focus. 
UMBRAE \cite{xia-umbrae-2024} employs a vision-language framework to decode fMRI signals, aligning textual descriptions with image features.

\subsection{Continual Learning}
We review two continual learning strategies \cite{wang-learningpromptcontinuallearning-2022}, each with distinct strengths and limitations. Rehearsal-based methods mitigate catastrophic forgetting by maintaining buffers of past task data \cite{hayes-memoryefficientexperiencereplay-2019, chaudhry-tinyepisodicmemoriescontinual-2019, chaudhry-usinghindsightanchorpast-2021, chaudhry-efficientlifelonglearningagem-2019, buzzega-darkexperiencegeneralcontinual-2020, rebuffi-icarlincrementalclassifierrepresentation-2017, reza-privacy-dl-2015, pham-dualnetcontinuallearningfast-2021, cha-co2lcontrastivecontinuallearning-2021, wu-largescaleincrementallearning-2019, wang2021continual, liu2025c, cai2023l2r, wan2022continual, cui2024learning, liu2024enhancing, wuavqacl, zhou2025ferret, wang2024self, yu2025language, hu2025kac, liu2025lora, yu2024addressing, kang2025your, wu2025bridging, tran2025boosting, resani2024miracle, serra2024federated, seo2024budgeted, peng2025tsvd, eskandar2025star, zhoubrainuicl, chenadaptive, woo2025meta, salami2024closed, ma2025vision, park2025active, farias2024self, zheng2025spurious, van2025computation}, allowing integration of old and new knowledge to prevent information loss. Some methods employ knowledge distillation to compress information across past and new tasks \cite{chaudhry-usinghindsightanchorpast-2021, buzzega-darkexperiencegeneralcontinual-2020, rebuffi-icarlincrementalclassifierrepresentation-2017, wu-largescaleincrementallearning-2019}, while others incorporate self-supervised learning techniques \cite{pham-dualnetcontinuallearningfast-2021, cha-co2lcontrastivecontinuallearning-2021, nguyen2019audio, nguyen2019sketch, nguyen2020self, nguyen2021clusformer, nguyen2022multi, nguyen2022two, nguyen2023algonauts, nguyen2023brainformer, nguyen2023fairness, nguyen2024bractive, nguyen2024diffusion, nguyen2024hierarchical, nguyen2024insect, nguyen2024qclusformer}.  Key challenges include buffer size constraints, which can degrade learning when buffers are excessively small \cite{cha-co2lcontrastivecontinuallearning-2021}, and data privacy concerns, which may limit access to stored data \cite{reza-privacy-dl-2015}. Architecture-based approaches counteract catastrophic forgetting by modifying the model structure.  This often involves introducing new parameter sets for each task \cite{zhao-deepbayesianunsupervisedlifelong-2021, yoon-lifelonglearningdynamicallyexpandable-2019, rusu-progressiveneuralnetworks-2022, rao-continualunsupervisedrepresentationlearning-2019, loo-generalizedvariationalcontinuallearning-2020, li-learngrowcontinualstructure-2019} or maintaining specialized sub-networks for different tasks \cite{wortsman-supermaskssuperposition-2020, serra-overcomingcatastrophicforgettinghard-2018, mallya-packnetaddingmultipletasks-2018, ke-continuallearningmixedsequence-2021}. 
Significant limitations of this method include increased model complexity due to additional parameters and the need to identify task types in advance to select the appropriate parameters, which is not always feasible during inference.

\begin{figure*}[!ht]
    \centering
    \includegraphics[width=1.0\linewidth]{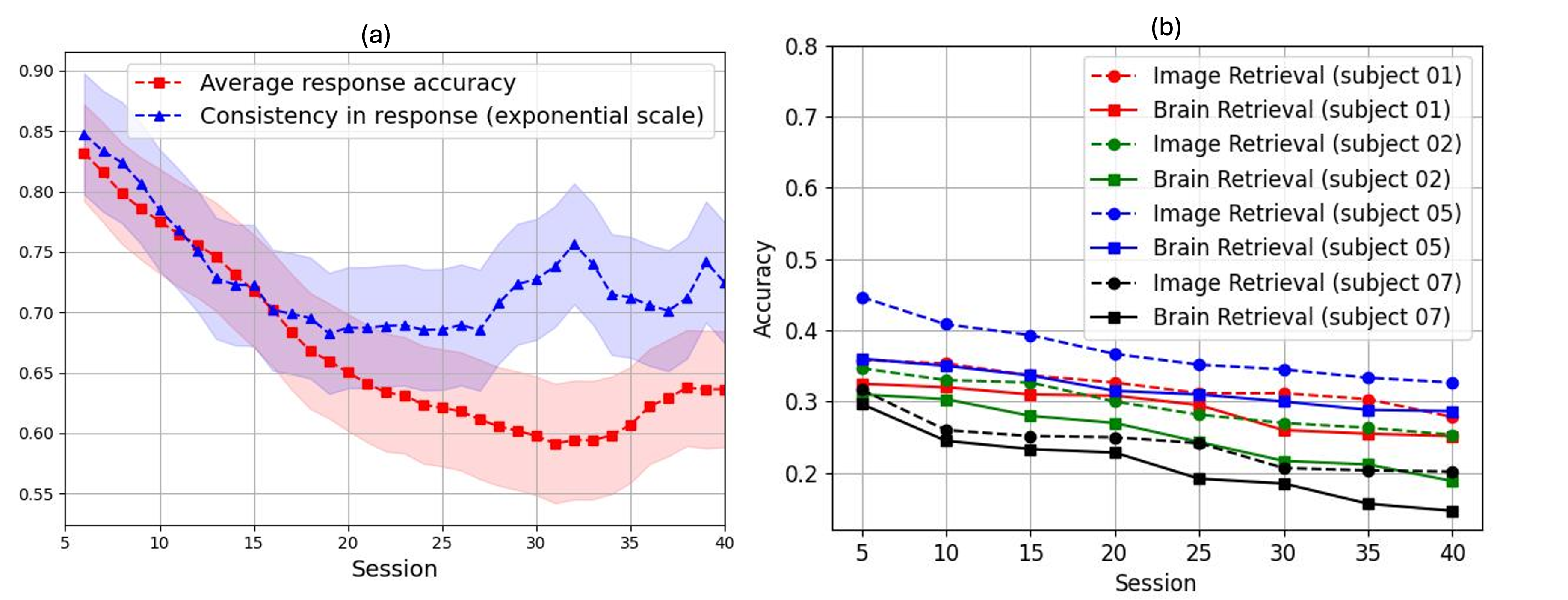}
    \caption{Data bias due to memory decay.  (a) Overall response accuracy \Bac{(range from 0 - 1) }and consistency in their response answers. (b) The retrieval performance of the model was trained every five sessions. \textbf{Best view in color}}
    \label{fig:statistic}
\end{figure*}

\section{Consistency of Brain Signals Analysis}
\label{sec:fmri_shift_drift}

In this section, we analyze the consistency of brain signals across sessions during data collection. First, we present statistical evidence highlighting inconsistencies in fMRI signals over sessions. Second, we demonstrate the impact of inconsistencies on recent vision-brain decoding models and discuss the importance of addressing this problem.

\subsection{Inconsistency in Brain Signals Over Time}
From the metadata of NSD, we obtain the response accuracy (Fig. \ref{fig:statistic} (a) - red line) of participants when determining whether they had seen the current visual stimulus in previous sessions (by pressing a button of Yes or No). Additionally, this data reveals whether participants are consistent with their answers by not changing their minds (Fig. \ref{fig:statistic} (a) - blue line), with only the final response being recorded.
Response accuracy declines over time, while the consistency of their answers also decreases. This pattern suggests growing uncertainty and declining confidence among participants as they experience memory decay.

\subsection{Impact of Inconsistency in Brain Signals on Models}

To investigate the impact of inconsistency on the brain signals, we utilize a vision-brain retrieval model \cite{mindeye2} trained \Bac{and validated} on data from every five sessions. \Bac{We keep the same model architecture and training parameters across training sessions.} The model is evaluated on two tasks: image retrieval, which involves identifying the corresponding image from a given brain signal, and brain retrieval, which involves finding the corresponding brain signal for a given image. As shown in Fig. \ref{fig:statistic} (b), the results reveal a decline in accuracy across sessions. This trend is likely due to the increasing inconsistency, which diminishes visual context information and makes learning more challenging for the model.

\section{The Proposed BRAIN Method}

In this section, we will first present our motivation, followed by the problem formulation. Then, we will introduce our proposed De-biased Contrastive Learning and Angular-based Forgetting Mitigation approaches to address the challenges in continual vision-brain understanding.

\subsection{Problem Motivation}

In Section \ref{sec:fmri_shift_drift}, we have highlighted a key challenge arising from the natural decay of human memory over time. As participants are repeatedly exposed to visual stimuli across multiple sessions, their ability to recognize previously seen images tends to decline in later sessions. This behavioral change introduces a critical issue: \textbf{\textit{bias in data representation}}, which poses a challenge for models in the Vision-Brain Understanding task. While prior studies have primarily overlooked this problem, addressing it is essential. This paper proposes a novel Continual Learning approach to tackle this issue for several reasons. First, as bias increases over time, i.e., across data collection sessions, a CL approach can adapt to these gradual changes, reducing the risk of outdated representations. Second, a CL approach aligns with the evolving nature of human memory and perception, making it a more biologically plausible solution than static training paradigms. Lastly, vision-brain data collection spans months or even years. Waiting until all data is collected before training would significantly delay research progress. The CL approach enables incremental model updates after each data collection session, facilitating continuous learning and timely adaptation. In the next section, we will detail the proposed Continual Learning approach to Vision-Brain Understanding.

\subsection{Problem Formulation}
In continual vision-brain understanding (CVBU), the goal is to learn an fMRI network ${F}$ on a sequence of data $\mathcal{D} = \{\mathcal{D}^1, \dots, \mathcal{D}^T\}$ where $T$ is the number of learning steps. At learning step $t$, the model ${F}$ encounters a dataset $\mathcal{D}_t = \{(x^t, y^t)\}$ where  $x^t \in \mathbb{R}^n$ is the fMRI signals length of $n$, and $y^t \in \mathbb{R}^{H \times W \times C}$ is corresponding visual stimuli. Here, we prefer $\mathcal{D}_t$ as the batch of data collected at session $t$. Let $z^t = F(x^t, \theta_t) \in \mathbb{R}^d$ be the fMRI features, $V$ be a frozen visual encoder that maps visual stimuli $y^t$ to visual centroid $c^t = V(y^t) \in \mathbb{R}^d$, and \Bac{$\theta_t$ is learnable parameters of $F$}. In the CVBU setup, the set of visual centroids $\mathcal{C}^t$ at step $t$ may include both centroids from previous learning steps and newly encountered ones. 
Formally, the CVBU model at step $t$ can be optimized as Eqn. \eqref{eqn:cl_objective}
\begin{align}
    \theta_t^* = \text{argmin } \mathbb{E}_{x^t, y^t \in \mathcal{D}^t} [\mathcal{L}_{C}(x^t, y^t) + \lambda_{CL}\mathcal{L}_{CL}(x^t)]
    \label{eqn:cl_objective}
\end{align}
where the contrastive loss $\mathcal{L}_C$ aligns $z^t$ with $c^t$, while the continual learning (CL) loss $\mathcal{L}_{CL}$ mitigates forgetting, weighted by the coefficient $\lambda_{CL}$. At learning step $t$, the model $F$ must align fMRI features with previously learned visual centroids and newly encountered visual centroids $\mathcal{C}^{t}$. This learning scenario presents two key challenges, i.e., Bias Data Representation and Catastrophic Forgetting. 
Therefore, we will present our approach to addressing these challenges in the following sections.

\subsection{De-bias Contrastive Learning}
\label{sec:ccl}
\begin{figure}[!b]
    \centering
    \includegraphics[width=\linewidth]{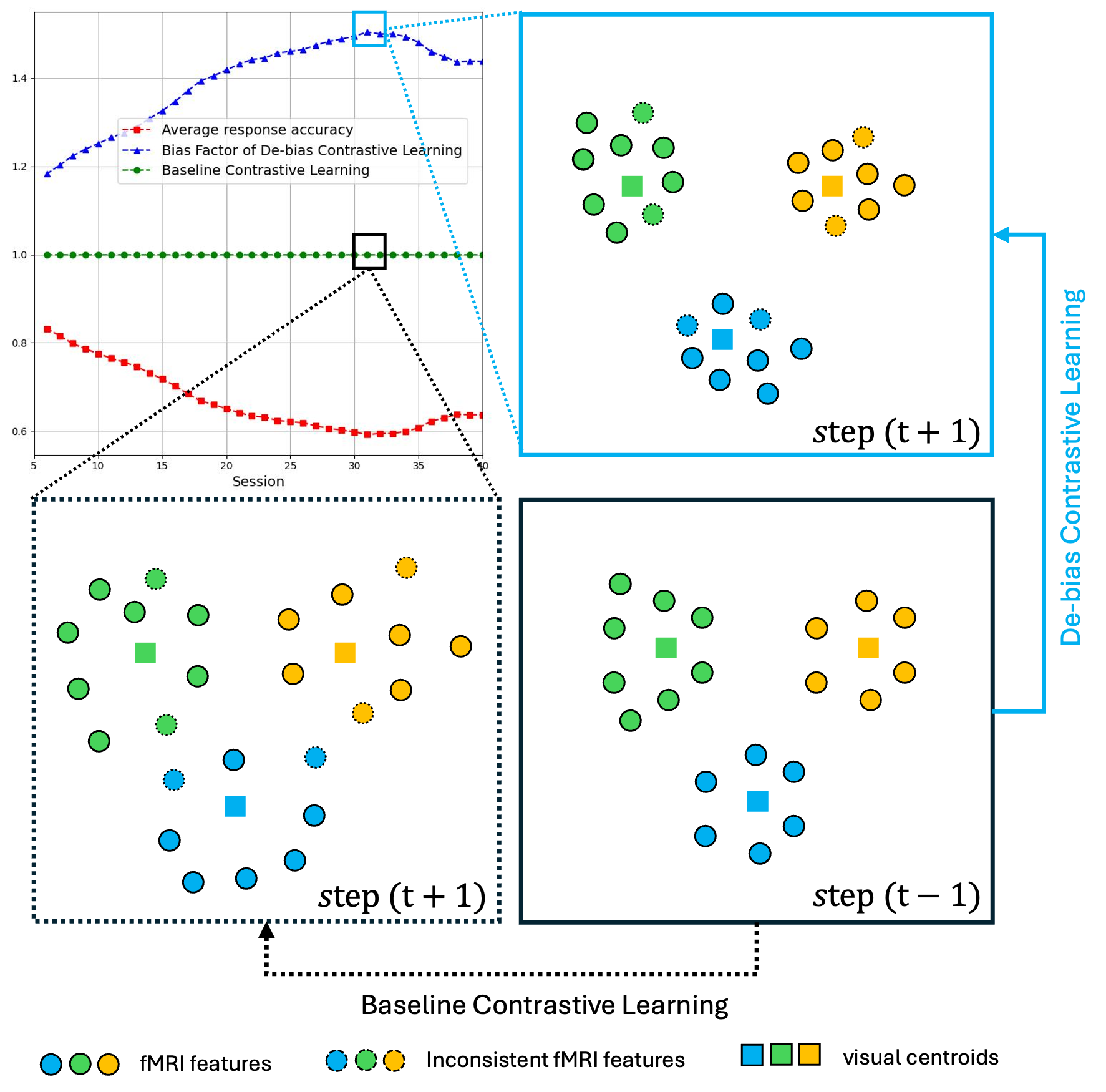}
    \caption{De-bias Contrastive Clustering.}
    \label{fig:ccl}
\end{figure}

\textbf{Limitations in Prior Methods.} 
Since the visual centroid $c^t$ is fixed using the frozen visual encoder $V$, the contrastive loss term in Eqn.~\eqref{eqn:cl_objective} can be rewritten as follows:  
\begin{equation}
\small
    \mathbb{E}_{x^t, y^t \in \mathcal{D}^t}\left[\mathcal{L}_{C}(x^t, y^t)\right] = \mathbb{E}_{x^t \sim p^t(x)}\left[\mathcal{L}_{C}(x^t, y^t)\right]
    \label{eqn:ccl_0}
\end{equation}  
where $p^t(x)$ represents the distribution of fMRI signals at step $t$.  Prior works \cite{mindeye1, mindeye2, mindbridge, xia-umbrae-2024} have assumed that $p^t(x) = p(x) \quad \forall t \leq T$ and ignored the bias factor.
In addition, prior studies \cite{truong2023falcon, cui2021parametric, cui2023generalized} have suggested that a biased data distribution can negatively influence contrastive learning, leading to inaccurate representations and scattered feature distributions (as shown in Figure \ref{fig:ccl}). To address this problem, we will model the bias factor in the next section and propose a novel loss to mitigate bias in vision-brain signals.

\noindent
\textbf{The Proposed De-bias Contrastive Learning (DCL).} 
As demonstrated in Section \ref{sec:fmri_shift_drift}, brain signals have bias in the later sessions, 
leading $p^t(x) \neq p^{t-1}(x)$. 
Consequently, Eqn. \eqref{eqn:ccl_0} can be further derived by the sampling technique as shown in Eqn. \eqref{eqn:ccl_1}.
\begin{equation}\label{eqn:ccl_1}
\small
\begin{split}
    \mathbb{E}_{x^t \in p^t(x)}\left[\mathcal{L}_{C}(x^t, y^t)\right] &= \mathbb{E}_{x^t \in p(x)}\left[\mathcal{L}_{C}(x^t, y^t) \frac{p^t(x)}{p(x)}\right] \\
    &= \mathbb{E}_{x^t \in p(x)}\left[\mathcal{L}_{C}(x^t, y^t) \frac{p(x|t)}{p(x)}\right] \\
    &= \mathbb{E}_{x^t \in p(x)}\left[\mathcal{L}_{C}(x^t, y^t) \frac{1}{p(t)}\right]    
\end{split}
\end{equation}
We call $w^t = \frac{1}{p(t)}$ as a \textit{bias} factor used to measure the difference between $p^t(x)$ and $p(x)$. To efficiently optimize Eqn. \eqref{eqn:ccl_1}, we need to model $ w^t $. We observe that $ w^t $ increases over time, indicating a growing bias in data representation. 
It suggests that $p(t)$ should be a decreasing function over time. Interestingly, in Section \ref{sec:fmri_shift_drift}, the bias increases due to a decrease in the average response accuracy of participants. Motivated by this observation, we propose a simple yet effective modeling of $ w^t $ as follows:  
\begin{equation}
    w^t = \frac{1}{p(t)} = e^{1 - r(t)}
\end{equation}
where $r(t)$ is the response accuracy at step $t$. The curve of $ w^t $ is illustrated in Fig. \ref{fig:ccl} (top left). According to this Figure, we infer that a higher weight should be assigned to the loss $ \mathcal{L}_{C} $ at step $ t $ if the participant experiences uncertainty or lack of confidence while processing the visual stimuli, and vice versa. The exponential function $e$ helps maintain smoothness in the bias factor and the gradient during optimization. In summary, we introduce De-bias Contrastive Learning, formulated in Eqn. \eqref{eqn:ccl_2}.
\begin{equation}\small
    \mathcal{L}_{C}(x^t, y^t) = -\text{log} \frac{\exp(z^t \times c^t)}{\sum_{z'} \exp(z' \times c^t)} e^{{1 - r(t)}}
    \label{eqn:ccl_2}
\end{equation}
\subsection{Angular-based Forgetting Mitigation}

\begin{figure}[!b]
    \centering
    \includegraphics[width=\linewidth]{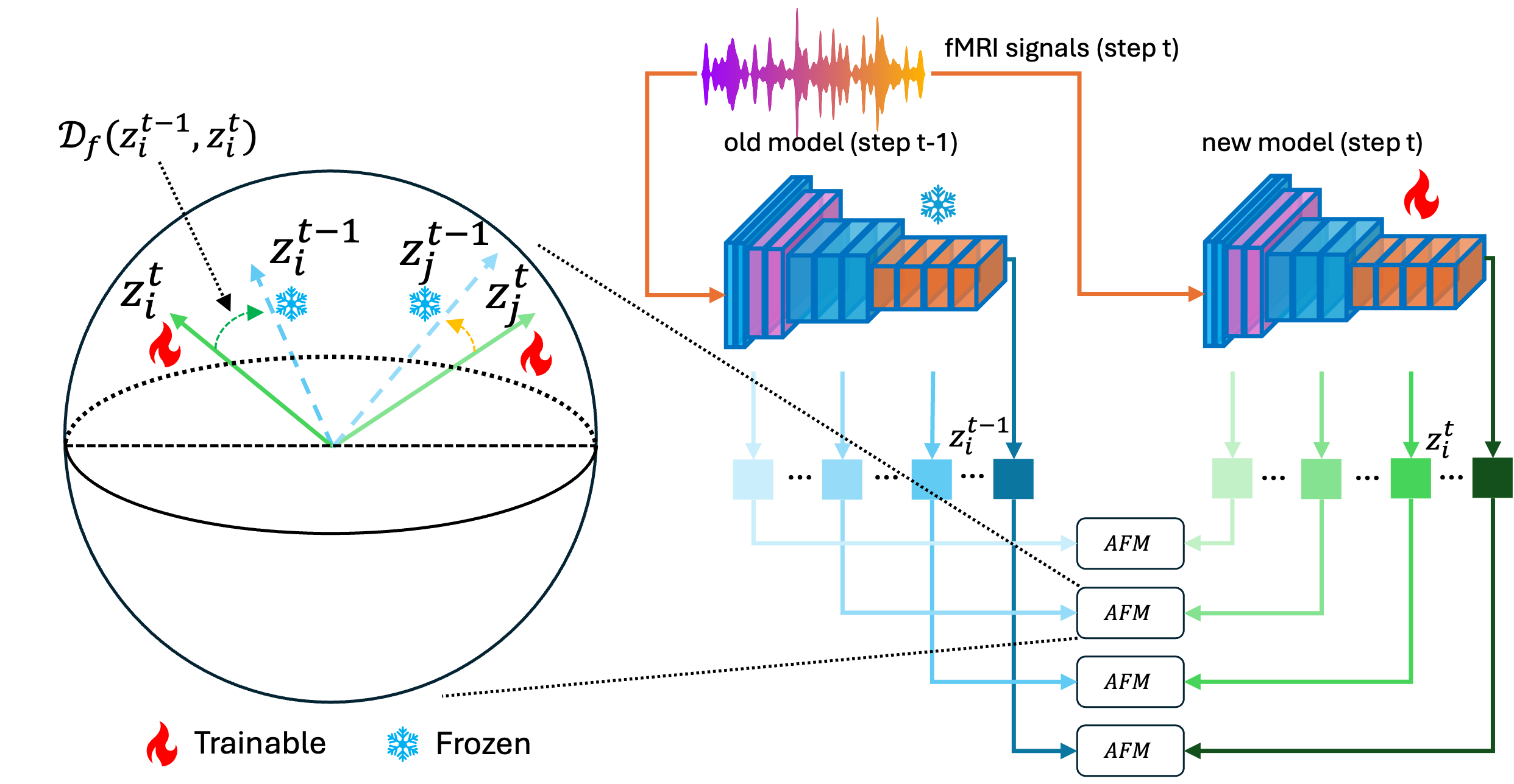}
    \caption{Angular-based Forgetting Mitigation.}
    \label{fig:distill}
\end{figure}

In continual vision-brain understanding, bias increases in the later learning steps, leading to catastrophic forgetting.
Moreover, this bias results in poor and uncertain representations of fMRI signals, making the model susceptible to overfitting on noise and leading to a significant drop in performance.
The forgetting problem is even more exaggerated during learning with the gradient descent algorithm, where the model parameters are updated globally. Then, the global changes in the model can significantly influence previously learned knowledge \cite{plop}.

To prevent the model from forgetting knowledge learned in previous sessions, prior work has adopted weight-regularization-based approaches \cite{kirkpatrick2017overcoming}. 
However, these methods remain limited in terms of flexibility and adaptability.
Indeed, regularization-based methods may impose overly strict constraints on weight updates, limiting the model's ability to adapt to new learning data. 
In addition, the model parameters learned from previous data may not always generalize well or adapt to distribution shifts in the current learning task due to the restrictive update constraints imposed by weight regularization.

Another approach to prevent catastrophic forgetting in continual learning is distillation, in which the model maintains its knowledge by imposing a constraint between the predictions of the old and new models \cite{plop, zhang2022representation}. 
Different from regulation-based methods \cite{kirkpatrick2017overcoming}, this approach finds a balance between being adaptable, where the model is enforced to maintain its old knowledge while allowing it to learn new data progressively, and being flexible, which allows the model to continuously update without forgetting.
Therefore, in our approach, we propose to model catastrophic forgetting using knowledge distillation.
Formally, our continual learning loss, $\mathcal{L}_{CL}$, in Eqn. \eqref{eqn:cl_objective}, can be reformed as:
\begin{equation}\label{eqn:ccl_3}
\small
\begin{split}
    \mathcal{L}_{CL}(x^t) = \mathcal{L}_{CL} (x^t, F, \theta_t, \theta_{t-1})
                          &= \frac{1}{L}\sum_{i=1}^L \mathcal{D}_f\left(F_i(x^t, \theta_{t-1}), F_i(x^t, \theta_{})\right) \\
                        &= \frac{1}{L}\sum_{i=1}^L \mathcal{D}_f\left(z^{t-1}_i, z^t_i\right)
\end{split}
\end{equation}
where $L$ is the number of intermediate features, $z^{t-1}_i = F_i(x^t, \theta_{t-1})$ and  $z^{t}_i = F_i(x^t, \theta_{t})$ are the $i^{th}$ intermediate feature extracted from model $F$ with $\theta_{t-1}$ and $\theta_{t}$, and $\mathcal{D}_f$ is the metric to measure the divergence between $z^{t-1}_i$ and  $z^t_i$. 
Fig. \ref{fig:distill} illustrates our knowledge distillation learning framework.

\noindent
\textbf{Limitations of Prior Knowledge Distillation.}
Prior studies commonly adopted the Euclidean distance ($\ell_2$) as a metric for $\mathcal{D}$ to measure the feature discrepancy.
However, using this metric in the continual learning paradigm could lead to limited performance.
In particular, using the constraint $\ell_2$ as a regularization to each $z^{t-1}_i$ and $z^t_i$, i.e. $\mathcal{D}_f\left(z^{t-1}_i, z^t_i\right) = || z_i^t - z_i^{t-1} ||^2_2$ (where $|| \cdot ||^2_2$ is the $\ell_2$ norm) could lead to the over-regularized problem. Although the $\ell_2$ metric has proven effective for the close-problem (i.e., non-incremental learning), it can be sensitive to feature variations in continual learning due to its reliance on absolute distances. When the model learns from new data, the scale of feature representations can change, leading to feature shifts that make it difficult to maintain knowledge learned in previous steps.

\noindent
\textbf{The Proposed Angular-based Forgetting Mitigation (AFM)}.In the continual learning setting, imposing feature discrepancy in the direction, i.e., an angular metric, rather than magnitude, is more valuable for two reasons. First, the angular metric is more robust against the scale and noise of feature representations \cite{deng2019arcface}. Second, since downstream tasks commonly use angular distances (e.g., fMRI-to-Image Retrieval \cite{mindeye1, mindeye2}, fMRI-to-Image Synthesis \cite{mindbridge, mindeye1, mindeye2, xia-umbrae-2024}) to compute similarity, using angular metrics could further improve performance in downstream applications. Therefore, \textit{the new angular metric is proposed to model the knowledge distillation} as: %
\begin{equation}
    \label{eqn:angular}
\footnotesize
    \mathcal{D}_f\left(z^{t-1}_i, z^t_i\right)
    = \left\| 1 - \frac{z^{t-1}_i}{||z^{t-1}_i||} * \frac{z^{t}_i}{||z^{t}_i||}\right\| ^2_2
\end{equation}
Under this form, our proposed distillation loss will retain knowledge of the model via the feature directions. Our learning approach provides the model with a degree of freedom to update its knowledge during the current learning step.

\subsection{Continual Vision-Brain Learning}

At each learning step, the vision-brain model $F$ is learned with the \textbf{De-bias Contrastive Loss} defined in Eqn. \eqref{eqn:ccl_1} and the \textbf{Angular-based Forgetting Mitigation} presented in Eqn. \eqref{eqn:angular}. 
Formally, the entire bias-mitigation continual learning objective in our approach can be defined as in Eqn.~\eqref{eqn:final-equation-css}.

\begin{equation}\label{eqn:final-equation-css}
\footnotesize
\begin{split}
    \mathcal{L} = \mathcal{L}_C(x^t, y^t) + \lambda_{CL}\mathcal{L}_{CL}(x^t)
                &= -\text{log} \frac{\exp(z^t \times c^t)}{\sum_{z'} \exp(z' \times c^t)} e^{1 - r(t)} \\ &
                +\lambda_{CL}\frac{1}{L} \sum\left\| 1 - \frac{z^{t-1}_i}{||z^{t-1}_i||} * \frac{z^{t}_i}{||z^{t}_i||}\right\| ^2_2
\end{split}
\end{equation}
\section{Datasets and Method Implementation}
In this section, we first detail the dataset used in our study and the evaluation protocol for assessing continual learning performance in Section \ref{subsec:dataset_evaluation_protocol}. The Natural Scenes Dataset (NSD) offers a unique opportunity to study fine-grained, session-wise brain activity across multiple participants, making it particularly well-suited for research on continual learning. Building on this dataset, we define an evaluation protocol that simulates real-world sequential training scenarios in which models must learn incrementally while retaining prior knowledge. This setup enables us to systematically investigate how different session splits and training strategies impact performance and memory retention. Next, we detail the implementation and training hyperparameters used in our experiments in Section \ref{subsec:implementation_details}

\subsection{Dataset and Evaluation Protocol}
\label{subsec:dataset_evaluation_protocol}
\textbf{Dataset.} We experiment on the Natural Scenes Dataset (NSD) \cite{nsd}. It was collected from eight participants across multiple sessions, in which each participant viewed a total of 73,000+ fMRI trials corresponding to 10,000 unique natural scene images sampled from the COCO dataset \cite{lin2014microsoft}. Each image was presented multiple times across different sessions, allowing researchers to analyze the reliability and consistency of brain responses. Each participant underwent extensive scanning (30-40 sessions per participant) while maintaining fixation on a central point. \Bac{To the best of our knowledge, the NSD dataset is the only publicly available resource that provides fine-grained data collection session information, which is essential for establishing a continual learning setup. This temporal structure enables the study of session-wise signal bias and its effect on brain-vision understanding, an aspect not supported by other existing datasets.}

\noindent
\textbf{Evaluation Protocol.} Let $(N_{init}, N_s)$ denote a pair representing a Continual Learning (CL) setup, where $N_{init}$ is the number of initial sessions used to train the initial model, and $N_s$ is the number of sessions incorporated in each subsequent training step. At each training step, the model is initialized with the pre-trained weights from the previous step, if available. During evaluation, only test samples of the previous and current sessions are used to measure the performance metric. Suppose we have $N_{session}$ data collection sessions. The total number of steps in the CL process is given by  $\frac{N_{session} - N_{init}}{N_s} + 1$. We experiment with four CL setups: $(20, 2)$, $(20, 5)$, $(20, 10)$, and $(15, 5)$.

\subsection{Implementation Details}
\label{subsec:implementation_details}
\textbf{fMRI Encoder and Vision Encoder}. We employ an fMRI encoder similar to \cite{mindeye1}. For the Vision Encoder $V$, we use a pre-trained visual encoder from the OpenCLIP models, such as ViT-B/16 \cite{clip-paper-2021}. For our distillation loss, we use features from the $L=3$ intermediate layers of the fMRI encoder.

\noindent
\textbf{Training Configuration}. Our framework is implemented in PyTorch and trained on one NVIDIA A100 GPU. The learning rate is initially set to $2.5e^{-4}$ and then gradually reduced to zero using CosineLinear \cite{martino2020semeval}. 
The CL weight factors $\lambda_{CL}$ are set to 1. The model is optimized by AdamW \cite{loshchilov2017decoupled} with a batch size of $16$ for $50$ epochs. The training is completed within two hours per subject.

\section{Experimental Results}

\begin{table*}[!t]
\caption{Average continual learning performance of $(\text{Brain} \rightarrow \text{Image})$ Retrieval task across different settings (15, 5), (20, 10), (20, 2) and (20, 5) on NSD.}
\label{tab:avg_cl_brain_image}
\resizebox{1.0\textwidth}{!}{

\begin{tabular}{|l|cccccc|ccc|ccccccccccc|ccccc|}

\hline

\multicolumn{1}{|l|}{\multirow{3}{*}{Method}} & \multicolumn{25}{c|}{\cellcolor[HTML]{CBCEFB}$(\text{Brain} \rightarrow \text{Image})$ Retrieval} \\ 

\cline{2-26} 

\multicolumn{1}{|l|}{} & \multicolumn{6}{c|}{\cellcolor[HTML]{FFCE93}(15, 5)} & \multicolumn{3}{c|}{\cellcolor[HTML]{FFCCC9}(20, 10)} & \multicolumn{11}{c|}{\cellcolor[HTML]{C0C0C0}(20, 2)} & \multicolumn{5}{c|}{\cellcolor[HTML]{DAE8FC}(20, 5)} \\ 

\cline{2-26} 

\multicolumn{1}{|l|}{} & \multicolumn{1}{c|}{\cellcolor[HTML]{FFCE93}1-15} & \multicolumn{1}{c|}{\cellcolor[HTML]{FFCE93}16-20} & \multicolumn{1}{c|}{\cellcolor[HTML]{FFCE93}21-25} & \multicolumn{1}{c|}{\cellcolor[HTML]{FFCE93}26-30} & \multicolumn{1}{c|}{\cellcolor[HTML]{FFCE93}31-35} & \multicolumn{1}{l|}{\cellcolor[HTML]{FFCE93}36-40} & \multicolumn{1}{c|}{\cellcolor[HTML]{FFCCC9}1-20} & \multicolumn{1}{c|}{\cellcolor[HTML]{FFCCC9}21-30} & \multicolumn{1}{c|}{\cellcolor[HTML]{FFCCC9}31-40} & \multicolumn{1}{c|}{\cellcolor[HTML]{C0C0C0}1-20} & \multicolumn{1}{c|}{\cellcolor[HTML]{C0C0C0}21-22} & \multicolumn{1}{c|}{\cellcolor[HTML]{C0C0C0}23-24} & \multicolumn{1}{c|}{\cellcolor[HTML]{C0C0C0}25-26} & \multicolumn{1}{c|}{\cellcolor[HTML]{C0C0C0}27-28} & \multicolumn{1}{c|}{\cellcolor[HTML]{C0C0C0}29-30} & \multicolumn{1}{c|}{\cellcolor[HTML]{C0C0C0}31-32} & \multicolumn{1}{c|}{\cellcolor[HTML]{C0C0C0}33-34} & \multicolumn{1}{c|}{\cellcolor[HTML]{C0C0C0}35-36} & \multicolumn{1}{c|}{\cellcolor[HTML]{C0C0C0}37-38} & \multicolumn{1}{c|}{\cellcolor[HTML]{C0C0C0}39-40} & \multicolumn{1}{c|}{\cellcolor[HTML]{DAE8FC}1-20} & \multicolumn{1}{c|}{\cellcolor[HTML]{DAE8FC}21-25} & \multicolumn{1}{c|}{\cellcolor[HTML]{DAE8FC}26-30} & \multicolumn{1}{c|}{\cellcolor[HTML]{DAE8FC}31-35} & \multicolumn{1}{l|}{\cellcolor[HTML]{DAE8FC}36-40} \\ 

\hline

W/o CL & 49.64 & 30.24 & 30.2 & 29.32 & 29.43 & 28.75 & 54.3 & 40.8 & 38.89 & 54.62 & 17.95 & 17.6 & 17.37 & 17.7 & 16.91 & 16.14 & 16.29 & 17.11 & 17.83 & 16.28 & 52.56 & 31.05 & 30.26 & 29.78 & 29.58 \\

LwF & 51.28 & 50.76 & 52.06 & 52.02 & 53.74 & 54.31 & 54.96 & 55.1 & 56.9 & 54.64 & 50.86 & 50.53 & 48.7 & 51.12 & 50.97 & 51.78 & 50.98 & 51.65 & 52.26 & 52.0 & 55.52 & 52.7 & 51.99 & 53.2 & 54.44 \\ 

PLOP & 53.24 & 52.19 & 53.19 & 53.62 & 55.42 & 56.56 & 56.95 & 57.1 & 58.48 & 56.52 & 52.3 & 52.8 & 51.16 & 52.06 & 52.4 & 52.94 & 52.64 & 53.94 & 54.44 & 53.97 & 56.47 & 54.27 & 54.46 & 55.54 & 56.44 \\

Ours & \textbf{55.97} & \textbf{54.88} & \textbf{56.78} & \textbf{56.87} & \textbf{57.86} & \textbf{58.98} & \textbf{59.51} & \textbf{59.32} & \textbf{62.19} & \textbf{59.11} & \textbf{55.11} & \textbf{55.3} & \textbf{53.97} & \textbf{55.55} & \textbf{55.3} & \textbf{56.25} & \textbf{55.94} & \textbf{57.15} & \textbf{57.56} & \textbf{56.96} & \textbf{59.74} & \textbf{57.81} & \textbf{56.54} & \textbf{58.42} & \textbf{58.71} \\

\hline

\end{tabular}%
}
\end{table*}

\begin{table*}[!t]
\caption{Average continual learning performance of $(\text{Image} \rightarrow \text{Brain})$ Retrieval task across different settings (15, 5), (20, 10), (20, 2), and (20, 5) on NSD.}
\label{tab:avg_cl_image_brain}
\resizebox{1.0\textwidth}{!}{

\begin{tabular}{|l|cccccc|ccc|ccccccccccc|ccccc|}

\hline

\multicolumn{1}{|l|}{\multirow{3}{*}{Method}} & \multicolumn{25}{c|}{\cellcolor[HTML]{FFFFC7}$(\text{Image} \rightarrow \text{Brain})$ Retrieval} \\ 

\cline{2-26} 

\multicolumn{1}{|l|}{} & \multicolumn{6}{c|}{\cellcolor[HTML]{FFCE93}(15, 5)} & \multicolumn{3}{c|}{\cellcolor[HTML]{FFCCC9}(20, 10)} & \multicolumn{11}{c|}{\cellcolor[HTML]{C0C0C0}(20, 2)} & \multicolumn{5}{c|}{\cellcolor[HTML]{DAE8FC}(20, 5)} \\ 

\cline{2-26} 

\multicolumn{1}{|l|}{} & \multicolumn{1}{c|}{\cellcolor[HTML]{FFCE93}1-15} & \multicolumn{1}{c|}{\cellcolor[HTML]{FFCE93}16-20} & \multicolumn{1}{c|}{\cellcolor[HTML]{FFCE93}21-25} & \multicolumn{1}{c|}{\cellcolor[HTML]{FFCE93}26-30} & \multicolumn{1}{c|}{\cellcolor[HTML]{FFCE93}31-35} & \multicolumn{1}{l|}{\cellcolor[HTML]{FFCE93}36-40} & \multicolumn{1}{c|}{\cellcolor[HTML]{FFCCC9}1-20} & \multicolumn{1}{c|}{\cellcolor[HTML]{FFCCC9}21-30} & \multicolumn{1}{c|}{\cellcolor[HTML]{FFCCC9}31-40} & \multicolumn{1}{c|}{\cellcolor[HTML]{C0C0C0}1-20} & \multicolumn{1}{c|}{\cellcolor[HTML]{C0C0C0}21-22} & \multicolumn{1}{c|}{\cellcolor[HTML]{C0C0C0}23-24} & \multicolumn{1}{c|}{\cellcolor[HTML]{C0C0C0}25-26} & \multicolumn{1}{c|}{\cellcolor[HTML]{C0C0C0}27-28} & \multicolumn{1}{c|}{\cellcolor[HTML]{C0C0C0}29-30} & \multicolumn{1}{c|}{\cellcolor[HTML]{C0C0C0}31-32} & \multicolumn{1}{c|}{\cellcolor[HTML]{C0C0C0}33-34} & \multicolumn{1}{c|}{\cellcolor[HTML]{C0C0C0}35-36} & \multicolumn{1}{c|}{\cellcolor[HTML]{C0C0C0}37-38} & \multicolumn{1}{c|}{\cellcolor[HTML]{C0C0C0}39-40} & \multicolumn{1}{c|}{\cellcolor[HTML]{DAE8FC}1-20} & \multicolumn{1}{c|}{\cellcolor[HTML]{DAE8FC}21-25} & \multicolumn{1}{c|}{\cellcolor[HTML]{DAE8FC}26-30} & \multicolumn{1}{c|}{\cellcolor[HTML]{DAE8FC}31-35} & \multicolumn{1}{l|}{\cellcolor[HTML]{DAE8FC}36-40} \\ 

\hline

W/o CL & 44.52 & 26.44 & 25.5 & 24.33 & 25.24 & 24.81 & 48.31 & 36.15 & 34.41 & 48.26 & 13.58 & 12.57 & 13.9 & 13.68 & 13.11 & 12.86 & 12.52 & 12.77 & 13.03 & 11.89 & 46.42 & 25.6 & 25.36 & 24.89 & 24.72 \\ 

LwF & 45.64 & 44.18 & 46.43 & 46.68 & 47.46 & 48.85 & 49.29 & 49.32 & 51.8 & 49.03 & 43.4 & 44.08 & 42.98 & 44.31 & 44.98 & 45.41 & 44.75 & 45.32 & 45.83 & 45.34 & 49.44 & 46.21 & 46.55 & 47.24 & 49.05 \\

PLOP & 47.04 & 46.2 & 47.27 & 48.61 & 49.37 & 51.05 & 50.61 & 51.17 & 53.08 & 50.92 & 45.44 & 45.04 & 45.18 & 46.06 & 45.76 & 46.87 & 46.06 & 47.3 & 48.03 & 47.65 & 51.33 & 47.8 & 49.0 & 49.48 & 50.88 \\

Ours & \textbf{50.64} & \textbf{49.25} & \textbf{50.44} & \textbf{51.51} & \textbf{51.97} & \textbf{53.62} & \textbf{54.61} & \textbf{53.95} & \textbf{56.72} & \textbf{53.72} & \textbf{48.37} & \textbf{49.6} & \textbf{48.52} & \textbf{48.66} & \textbf{49.27} & \textbf{50.39} & \textbf{49.23} & \textbf{51.15} & \textbf{50.85} & \textbf{50.64} & \textbf{53.97} & \textbf{51.37} & \textbf{51.02} & \textbf{52.55} & \textbf{54.06} \\

\hline
\end{tabular}%
}
\end{table*}

\begin{table*}[!t]
\caption{Ablation studies on the effectiveness of the proposed approach on $(\text{Brain} \rightarrow \text{Image})$ Retrieval task.  Non-CL: Non Continual Learning. Contras: Contrastive Learning, RA: Response Accuracy, BA: Brain Activity, DCL: De-bias Contrastive Learning, AFM: Angular-based Forgetting Mitigation}
\label{tab:abl_brain_image}

\resizebox{1.0\textwidth}{!}{

\begin{tabular}{|l|l|l|l|l|ccccccccccccccccccccccccc|}
\hline

\multicolumn{1}{|l|}{} & \multicolumn{1}{l|}{} & \multicolumn{1}{l|}{} & \multicolumn{1}{l|}{} & \multicolumn{1}{l|}{} & \multicolumn{25}{c|}{\cellcolor[HTML]{CBCEFB}$(\text{Brain} \rightarrow \text{Image})$ Retrieval} \\ 

\cline{6-30} 

\multicolumn{1}{|l|}{} & \multicolumn{1}{l|}{} & \multicolumn{1}{l|}{} & \multicolumn{1}{l|}{} & \multicolumn{1}{l|}{} & \multicolumn{6}{c|}{\cellcolor[HTML]{FFCE93}(15, 5)} & \multicolumn{3}{c|}{\cellcolor[HTML]{FFCCC9}(20, 10)} & \multicolumn{11}{c|}{\cellcolor[HTML]{C0C0C0}(20, 2)} & \multicolumn{5}{c|}{\cellcolor[HTML]{DAE8FC}(20, 5)} \\ 

\cline{6-30} 

\multicolumn{1}{|l|}{\multirow{-3}{*}{Exp}} & \multicolumn{1}{l|}{\multirow{-3}{*}{Method}} & \multicolumn{1}{l|}{\multirow{-3}{*}{Alignment}} & \multicolumn{1}{l|}{\multirow{-3}{*}{Forgetting}} & \multicolumn{1}{l|}{\multirow{-3}{*}{Rehearsal}} & \multicolumn{1}{c|}{\cellcolor[HTML]{FFCE93}1-15} & \multicolumn{1}{c|}{\cellcolor[HTML]{FFCE93}16-20} & \multicolumn{1}{c|}{\cellcolor[HTML]{FFCE93}21-25} & \multicolumn{1}{c|}{\cellcolor[HTML]{FFCE93}26-30} & \multicolumn{1}{c|}{\cellcolor[HTML]{FFCE93}31-35} & \multicolumn{1}{l|}{\cellcolor[HTML]{FFCE93}36-40} & \multicolumn{1}{c|}{\cellcolor[HTML]{FFCCC9}1-20} & \multicolumn{1}{c|}{\cellcolor[HTML]{FFCCC9}21-30} & \multicolumn{1}{c|}{\cellcolor[HTML]{FFCCC9}31-40} & \multicolumn{1}{c|}{\cellcolor[HTML]{C0C0C0}1-20} & \multicolumn{1}{c|}{\cellcolor[HTML]{C0C0C0}21-22} & \multicolumn{1}{c|}{\cellcolor[HTML]{C0C0C0}23-24} & \multicolumn{1}{c|}{\cellcolor[HTML]{C0C0C0}25-26} & \multicolumn{1}{c|}{\cellcolor[HTML]{C0C0C0}27-28} & \multicolumn{1}{c|}{\cellcolor[HTML]{C0C0C0}29-30} & \multicolumn{1}{c|}{\cellcolor[HTML]{C0C0C0}31-32} & \multicolumn{1}{c|}{\cellcolor[HTML]{C0C0C0}33-34} & \multicolumn{1}{c|}{\cellcolor[HTML]{C0C0C0}35-36} & \multicolumn{1}{c|}{\cellcolor[HTML]{C0C0C0}37-38} & \multicolumn{1}{c|}{\cellcolor[HTML]{C0C0C0}39-40} & \multicolumn{1}{c|}{\cellcolor[HTML]{DAE8FC}1-20} & \multicolumn{1}{c|}{\cellcolor[HTML]{DAE8FC}21-25} & \multicolumn{1}{c|}{\cellcolor[HTML]{DAE8FC}26-30} & \multicolumn{1}{c|}{\cellcolor[HTML]{DAE8FC}31-35} & \multicolumn{1}{l|}{\cellcolor[HTML]{DAE8FC}36-40} \\ 

\hline

1 & Non-CL & Contras & \xmark & \xmark & \textcolor{gray}{55.76} & \textcolor{gray}{54.79} & \textcolor{gray}{56.32} & \textcolor{gray}{57.83} & \textcolor{gray}{57.96} & \textcolor{gray}{59.81} & \textcolor{gray}{59.4} & \textcolor{gray}{58.92} & \textcolor{gray}{60.29} & \textcolor{gray}{58.82} & \textcolor{gray}{56.76} & \textcolor{gray}{54.89} & \textcolor{gray}{53.97} & \textcolor{gray}{55.62} & \textcolor{gray}{55.44} & \textcolor{gray}{55.65} & \textcolor{gray}{56.28} & \textcolor{gray}{56.14} & \textcolor{gray}{57.76} & \textcolor{gray}{57.51} & \textcolor{gray}{59.45} & \textcolor{gray}{57.58} & \textcolor{gray}{57.49} & \textcolor{gray}{58.5} & \textcolor{gray}{59.42} \\

2 & Baseline & Contras & $\ell_2$ & \xmark & 51.52 & 50.63 & 52.11 & 52.84 & 53.32 & 54.42 & 54.81 & 54.3 & 55.98 & 54.77 & 52.29 & 50.38 & 49.73 & 51.09 & 50.8 & 50.74 & 51.96 & 51.31 & 52.62 & 52.32 & 55.38 & 52.94 & 53.28 & 53.76 & 54.62 \\

3 & Ours & DCL + RA & $\ell_2$ & \xmark & 53.24 & 52.38 & 54.6 & 54.6 & 55.91 & 57.43 & 57.13 & 56.89 & 59.04 & 57.39 & 53.17 & 52.83 & 52.08 & 53.71 & 53.44 & 53.77 & 53.54 & 54.46 & 54.6 & 54.15 & 57.8 & 55.53 & 55.01 & 55.73 & 57.7 \\

4 & Ours & DCL + BA & AFM & \xmark & 54.56 & 53.37 & 54.5 & 55.77 & 55.61 & 57.2 & 57.66 & 56.81 & 58.63 & 57.68 & 54.04 & 53.25 & 52.35 & 53.39 & 53.84 & 54.04 & 54.34 & 53.88 & 55.31 & 54.52 & 58.22 & 55.8 & 56.0 & 56.62 & 57.65 \\ 

5 & Ours & DCL + RA & AFM & \cmark & 49.51 & 48.84 & 50.31 & 51.76 & 51.76 & 53.05 & 53.61 & 52.83 & 54.18 & 53.0 & 50.81 & 48.52 & 48.87 & 49.17 & 48.69 & 49.32 & 50.37 & 50.11 & 51.63 & 51.08 & 53.28 & 50.97 & 51.36 & 52.73 & 53.52 \\

6 & Ours & DCL + RA& AFM & \xmark & \textbf{55.97} & \textbf{54.88} & \textbf{56.78} & \textbf{56.87} & \textbf{57.86} & \textbf{58.98} & \textbf{59.51} & \textbf{59.32} & \textbf{62.19} & \textbf{59.11} & \textbf{55.11} & \textbf{55.3} & \textbf{53.97} & \textbf{55.55} & \textbf{55.3} & \textbf{56.25} & \textbf{55.94} & \textbf{57.15} & \textbf{57.56} & \textbf{56.96} & \textbf{59.74} & \textbf{57.81} & \textbf{56.54} & \textbf{58.42} & \textbf{58.71} \\

\hline

\end{tabular}
}
\end{table*}

\begin{table*}[!t]
\caption{Ablation studies on the effectiveness of the proposed approach on 
$(\text{Image} \rightarrow \text{Brain})$ Retrieval task.  Non-CL: Non Continual Learning. Contras: Contrastive Learning, RA: Response Accuracy, BA: Brain Activity, DCL: De-bias Contrastive Learning, AFM: Angular-based Forgetting Mitigation}
\label{tab:abl_image_brain}

\resizebox{1.0\textwidth}{!}{

\begin{tabular}{|l|l|l|l|l|ccccccccccccccccccccccccc|}
\hline

\multicolumn{1}{|l|}{} & \multicolumn{1}{l|}{} & \multicolumn{1}{l|}{} & \multicolumn{1}{l|}{} & \multicolumn{1}{l|}{} & \multicolumn{25}{c|}{\cellcolor[HTML]{FFFFC7}$(\text{Image} \rightarrow \text{Brain})$ Retrieval} \\ 

\cline{6-30} 

\multicolumn{1}{|l|}{} & \multicolumn{1}{l|}{} & \multicolumn{1}{l|}{} & \multicolumn{1}{l|}{} & \multicolumn{1}{l|}{} & \multicolumn{6}{c|}{\cellcolor[HTML]{FFCE93}(15, 5)} & \multicolumn{3}{c|}{\cellcolor[HTML]{FFCCC9}(20, 10)} & \multicolumn{11}{c|}{\cellcolor[HTML]{C0C0C0}(20, 2)} & \multicolumn{5}{c|}{\cellcolor[HTML]{DAE8FC}(20, 5)} \\ 

\cline{6-30} 

\multicolumn{1}{|l|}{\multirow{-3}{*}{Exp}} & \multicolumn{1}{l|}{\multirow{-3}{*}{Method}} & \multicolumn{1}{l|}{\multirow{-3}{*}{Alignment}} & \multicolumn{1}{l|}{\multirow{-3}{*}{Forgetting}} & \multicolumn{1}{l|}{\multirow{-3}{*}{Rehearsal}} & \multicolumn{1}{c|}{\cellcolor[HTML]{FFCE93}1-15} & \multicolumn{1}{c|}{\cellcolor[HTML]{FFCE93}16-20} & \multicolumn{1}{c|}{\cellcolor[HTML]{FFCE93}21-25} & \multicolumn{1}{c|}{\cellcolor[HTML]{FFCE93}26-30} & \multicolumn{1}{c|}{\cellcolor[HTML]{FFCE93}31-35} & \multicolumn{1}{l|}{\cellcolor[HTML]{FFCE93}36-40} & \multicolumn{1}{c|}{\cellcolor[HTML]{FFCCC9}1-20} & \multicolumn{1}{c|}{\cellcolor[HTML]{FFCCC9}21-30} & \multicolumn{1}{c|}{\cellcolor[HTML]{FFCCC9}31-40} & \multicolumn{1}{c|}{\cellcolor[HTML]{C0C0C0}1-20} & \multicolumn{1}{c|}{\cellcolor[HTML]{C0C0C0}21-22} & \multicolumn{1}{c|}{\cellcolor[HTML]{C0C0C0}23-24} & \multicolumn{1}{c|}{\cellcolor[HTML]{C0C0C0}25-26} & \multicolumn{1}{c|}{\cellcolor[HTML]{C0C0C0}27-28} & \multicolumn{1}{c|}{\cellcolor[HTML]{C0C0C0}29-30} & \multicolumn{1}{c|}{\cellcolor[HTML]{C0C0C0}31-32} & \multicolumn{1}{c|}{\cellcolor[HTML]{C0C0C0}33-34} & \multicolumn{1}{c|}{\cellcolor[HTML]{C0C0C0}35-36} & \multicolumn{1}{c|}{\cellcolor[HTML]{C0C0C0}37-38} & \multicolumn{1}{c|}{\cellcolor[HTML]{C0C0C0}39-40} & \multicolumn{1}{c|}{\cellcolor[HTML]{DAE8FC}1-20} & \multicolumn{1}{c|}{\cellcolor[HTML]{DAE8FC}21-25} & \multicolumn{1}{c|}{\cellcolor[HTML]{DAE8FC}26-30} & \multicolumn{1}{c|}{\cellcolor[HTML]{DAE8FC}31-35} & \multicolumn{1}{l|}{\cellcolor[HTML]{DAE8FC}36-40} \\ 

\hline

1 & Non-CL & Contras & \xmark & \xmark & \textcolor{gray}{50.96} & \textcolor{gray}{48.96} & \textcolor{gray}{50.7} & \textcolor{gray}{51.07} & \textcolor{gray}{52.36} & \textcolor{gray}{53.36} & \textcolor{gray}{54.0} & \textcolor{gray}{53.4} & \textcolor{gray}{55.72} & \textcolor{gray}{54.28} & \textcolor{gray}{49.74} & \textcolor{gray}{46.86} & \textcolor{gray}{46.58} & \textcolor{gray}{48.8} & \textcolor{gray}{48.82} & \textcolor{gray}{49.6} & \textcolor{gray}{49.11} & \textcolor{gray}{48.95} & \textcolor{gray}{50.84} & \textcolor{gray}{49.97} & \textcolor{gray}{53.8} & \textcolor{gray}{51.62} & \textcolor{gray}{52.14} & \textcolor{gray}{53.83} & \textcolor{gray}{53.54} \\ 

2 & Baseline & Contras & $\ell_2$ & \xmark & 46.91 & 43.65 & 46.21 & 46.64 & 47.76 & 48.61 & 48.76 & 48.78 & 51.49 & 49.46 & 46.03 & 43.14 & 42.55 & 44.02 & 44.7 & 44.02 & 44.44 & 43.96 & 45.55 & 46.01 & 48.79 & 47.03 & 47.82 & 48.05 & 48.93 \\

3 & Ours & DCL + RA & $\ell_2$ & \xmark & 47.79 & 46.37 & 48.38 & 49.48 & 50.74 & 52.28 & 51.75 & 51.96 & 54.4 & 51.77 & 46.66 & 46.38 & 45.84 & 46.59 & 47.14 & 47.6 & 47.41 & 47.8 & 48.78 & 48.39 & 51.38 & 49.36 & 49.24 & 50.28 & 51.85 \\

4 & Ours & DCL + BA & AFM & \xmark & 48.99 & 47.34 & 48.53 & 49.76 & 50.04 & 50.42 & 51.4 & 51.77 & 53.76 & 52.06 & 47.58 & 45.41 & 44.66 & 46.51 & 47.04 & 47.12 & 47.66 & 47.14 & 48.14 & 48.45 & 52.12 & 49.53 & 50.3 & 51.04 & 52.15 \\

5 & Ours & DCL + RA & AFM & \cmark & 44.3 & 42.44 & 45.1 & 45.07 & 46.03 & 46.44 & 47.49 & 47.13 & 49.68 & 47.04 & 44.32 & 41.5 & 40.3 & 42.53 & 42.18 & 43.02 & 42.85 & 42.36 & 44.23 & 44.31 & 47.26 & 45.42 & 45.31 & 47.16 & 47.68 \\

6 & Ours & DCL + RA& AFM & \xmark & \textbf{50.64} & \textbf{49.25} & \textbf{50.44} & \textbf{51.51} & \textbf{51.97} & \textbf{53.62} & \textbf{54.61} & \textbf{53.95} & \textbf{56.72} & \textbf{53.72} & \textbf{48.37} & \textbf{49.6} & \textbf{48.52} & \textbf{48.66} & \textbf{49.27} & \textbf{50.39} & \textbf{49.23} & \textbf{51.15} & \textbf{50.85} & \textbf{50.64} & \textbf{53.97} & \textbf{51.37} & \textbf{51.02} & \textbf{52.55} & \textbf{54.06} \\

\hline

\end{tabular}
}
\end{table*}

In this section, we first discuss the experimental results for average continual learning performance, as well as individual performance across different continual learning setups. We also conduct comprehensive ablation studies to analyze 

\subsection{Continual Learning Results}
\begin{figure*}[!t]
    \centering
    \includegraphics[width=1.0\linewidth]{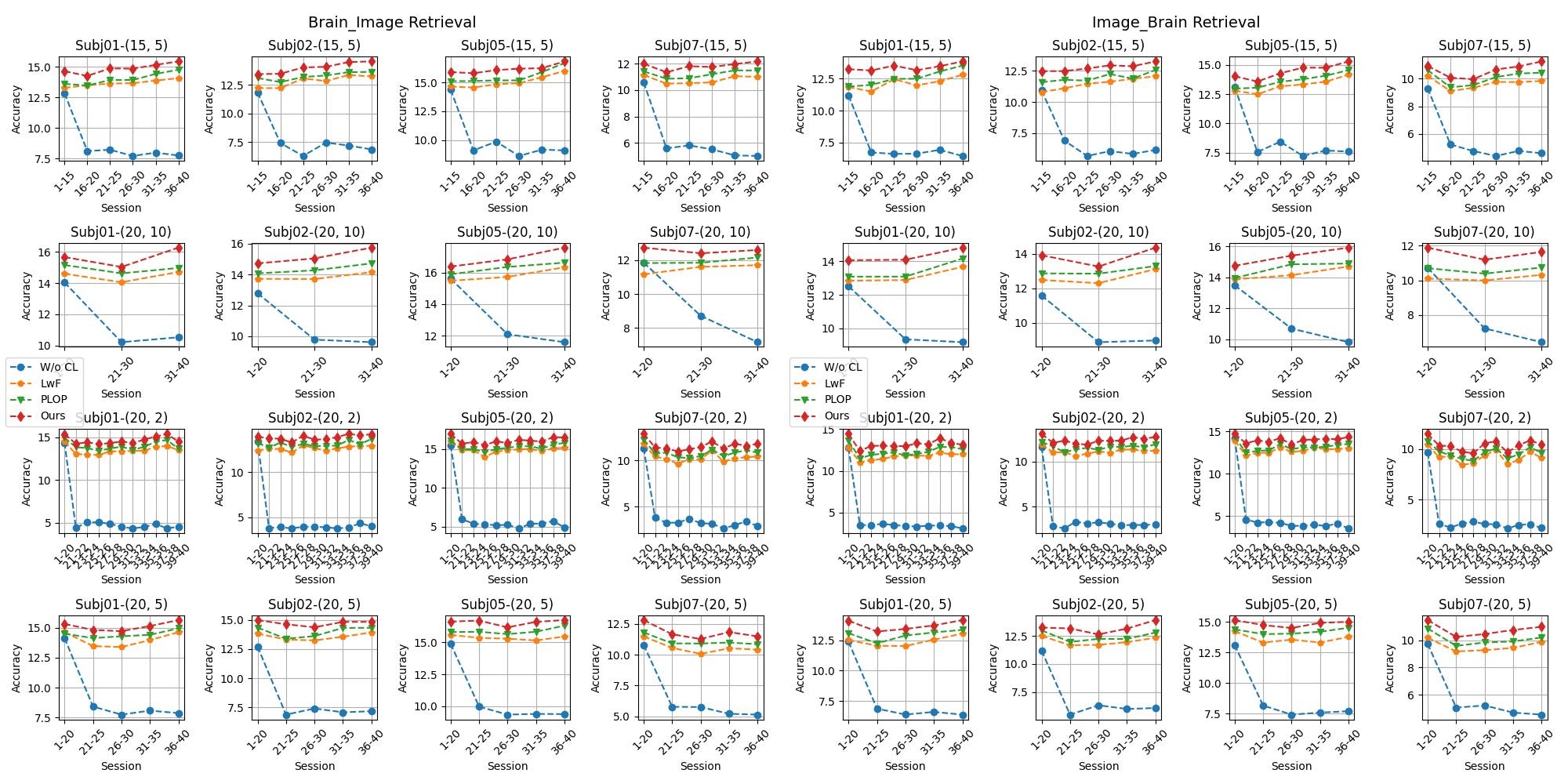}
    \caption{Details continual learning performance of all subjects (subj01, subj02, subj05, and subj07) across different settings  $(15, 5), (20, 10), (20, 2) \text{ and } (20, 5)$ on NSD.}
    \label{fig:cl}
\end{figure*}

Table \ref{tab:avg_cl_brain_image} and Table \ref{tab:avg_cl_image_brain} provide an average continual learning performance, and Fig. \ref{fig:cl} details the performance on four subjects across different training settings on NSD. These results compare standard training without continual learning (W/o CL), Learning without Forgetting (LwF \cite{lwf}), PLOP \cite{plop}, and our proposed method under various incremental update configurations: $(15, 5), (20, 10), (20, 2) \text{ and } (20, 5)$. Across all settings, W/o CL exhibits a sharp decline in performance, confirming severe forgetting when learning new session data. For instance, in subject 01 under $(20, 2)$, W/o CL achieves 61.57\% at the first step ($1-20$), but 18.06\% ($\downarrow$ 43.57\%) at the end $(39-40)$. Meanwhile, our proposed method achieves 57.97\% ($\downarrow$ 3.6\%). LwF and PLOP mitigate forgetting to some extent, but their effectiveness varies across settings. In particular, under the $(20, 2)$ setting, LwF achieves 54.7\%, outperforming PLOP at 54.14\%, but still lags behind our method at 57.97\%. Our method consistently maintains higher accuracy in various subjects and learning settings, suggesting a more stable learning mechanism.

\subsection{Ablation Studies}
\begin{figure*}[!tb]
    \centering
    \includegraphics[width=1.0\linewidth]{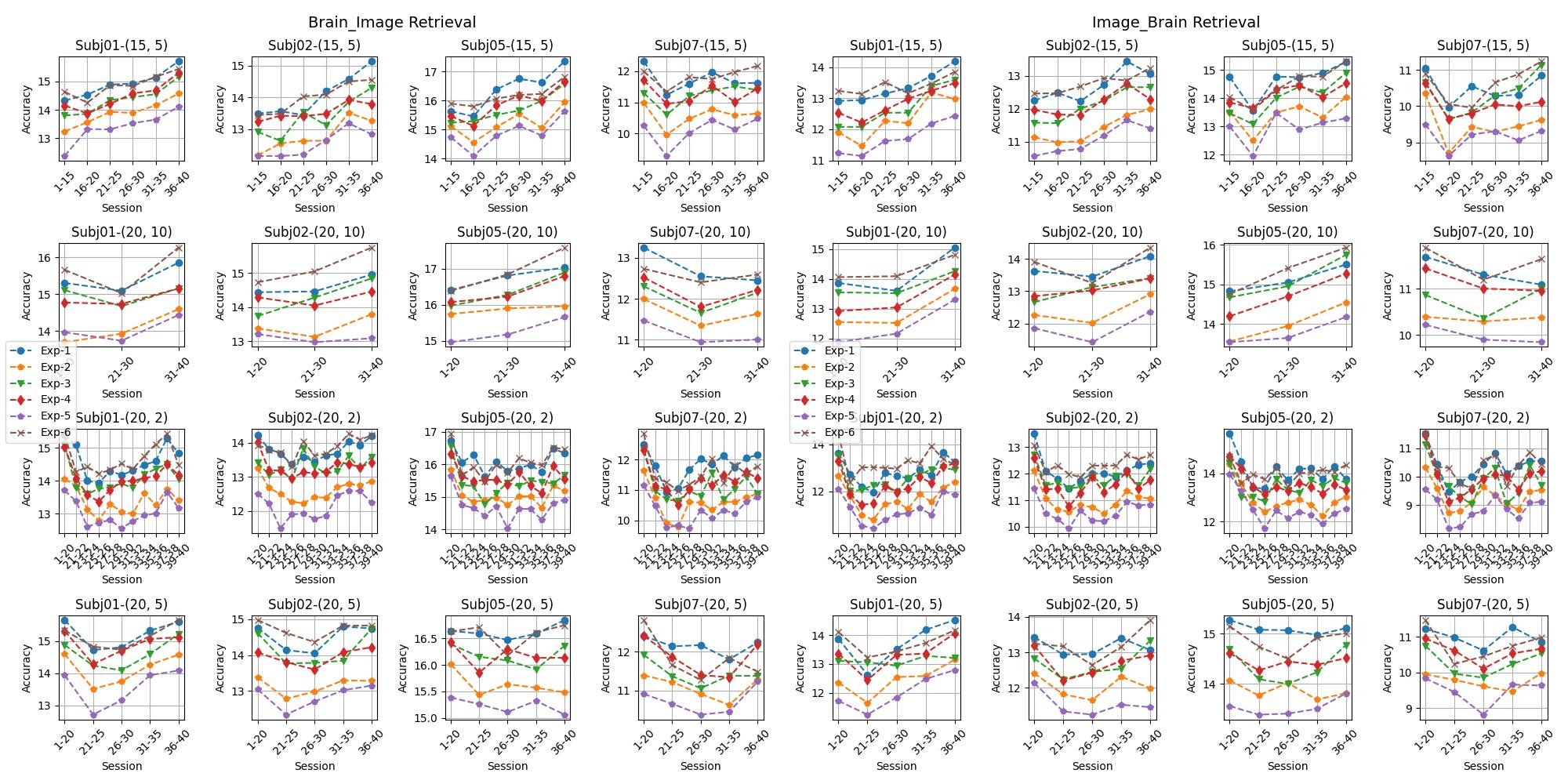}
    \caption{Details ablation study performance of all subjects (subj01, subj02, subj05, and subj07) across different settings  $(15, 5), (20, 10), (20, 2) \text{ and } (20, 5)$ on NSD. The detail settings of $\text{Exps} 1-6$
    can be found in Table \ref{tab:abl_brain_image} and Table \ref{tab:abl_image_brain}}
    \label{fig:abl}
\end{figure*}

We report the average performance across all subjects in Table \ref{tab:abl_brain_image} and Table \ref{tab:abl_image_brain}, while Fig. \ref{fig:abl} illustrates the performance details of each subject.

\noindent
\textbf{Effectiveness of De-bias Contrastive Learning.}  
We evaluate the effectiveness of DCL by comparing it with the baseline that uses vanilla Contrastive Learning. Here, DCL employs Response Accuracy (RA) to model the bias. To mitigate catastrophic forgetting, we incorporate $\ell_2$ regularization. As shown in Table \ref{tab:abl_brain_image} and Table \ref{tab:abl_image_brain}, the model trained with $(\text{DCL + RA},  \ell_2)$ in $(\text{Exp}-3)$ 
 outperforms the model trained with $(\text{Contras}, \ell_2)$ in $(\text{Exp}-2)$ by 1\%-4\% across various subjects (01, 02, 05, 07), continual learning setups \((15, 5), (20, 10), (20, 2), (20, 5)\), and retrieval tasks \((\text{Brain} \rightarrow \text{Image}), (\text{Image} \rightarrow \text{Brain})\). These results demonstrate the superior performance of DCL over typical Contrastive Learning.
 
\noindent
\textbf{Effectiveness of Angular-based Forgetting Mitigation.}  
We further assess the impact of AFM compared to $\ell_2$ regularization. In this experiment, we use $(\text{DCL + RA})$ for alignment. As presented in Table \ref{tab:abl_brain_image} and Table \ref{tab:abl_image_brain}, the model trained with $(\text{DCL + RA},  \text{AFM})$ in $(\text{Exp}-6)$ achieves 1\%-3\% higher performance than $(\text{DCL + RA},  \ell_2)$ in $(\text{Exp}-3)$ across the same subjects, continual learning setups, and retrieval tasks. This improvement suggests that AFM is more robust to noise than $\ell_2$, making it better suited for Vision-Brain Understanding tasks.

\noindent
\textbf{Non-Continual Learning Setting.}  In the Table \ref{tab:abl_brain_image} and Table \ref{tab:abl_image_brain},
we compare our continual learning approach in $(\text{Exp}-6)$ to the Non-Continual Learning (Non-CL) setup in $(\text{Exp}-1)$, where data from all sessions is combined and trained in a single stage. The results show that our approach achieves competitive performance, surpassing the Non-CL setup—an uncommon outcome in other domains.

\noindent
\textbf{Modeling the Bias Factor.}  In Section \ref{sec:ccl}, we proposed modeling the bias factor $w^t$ using participants' accuracy responses during data collection, which tend to increase over sessions. Here, we explore an alternative approach to modeling this factor.  Specifically, as participants experience uncertainty, their brain activity weakens, decreasing the number of activated voxels—i.e., voxels with values above zero. Motivated by this observation, we propose modeling $w^t = \frac{1}{p(t)}$ as: $w^t = e^{1-a(t)} = e^{1 - N_{a} / N_b}$ where $N_a$ represents the number of activated brain voxels, and $ N_b $ is the total number of brain voxels. Performance results are reported in Table \ref{tab:abl_brain_image} and Table \ref{tab:abl_image_brain}. Overall, modeling the bias factor using Brain Activation ($\text{Exp} - 4$) yields slightly lower performance than the Accuracy Response ($\text{Exp} - 6$) approach. However, in practice, Accuracy Response may not always be available in certain datasets, whereas brain activity data is more accessible. Furthermore, Brain Activation-based modeling still outperforms previous methods such as LwF \cite{lwf} and PLOP \cite{plop}. Therefore, in some cases, Brain Activity can serve as a viable alternative for modeling consistency in the brain.

\noindent
\textbf{Rehearsal-Free and Rehearsal-Based Approaches.}   In this section, we evaluate our proposed continual learning under a rehearsal-based approach. Specifically, after each training step $t$, we randomly retain 10\% of the data and reuse it in the next step. Notably, we do not apply AFM in the rehearsal-based approach. The performance results are reported in Table \ref{tab:abl_brain_image} and Table \ref{tab:abl_image_brain}.  Our findings indicate that the rehearsal-based approach ($\text{Exp} - 5$)  performs significantly worse than the rehearsal-free approach ($\text{Exp} - 6$)  across various settings. It is because each sample pair forms an individual cluster, and retaining only 10\% of the data prevents the loss of knowledge for those specific samples while potentially leading to the loss of information from the remaining 90\%. For this reason, the rehearsal-based method is not an ideal solution for continual vision-brain understanding.

\begin{table}[!t]
\centering
\caption{Brain-Image retrieval results on the NSD.}
\label{tab:image_brain_retrieval}
\small %
\begin{tabular}{lcc}
\textbf{Method} & \textbf{Image-Brain}$\uparrow$ & \textbf{Brain-Image}$\uparrow$ \\
\hline
Lin et al. \cite{lin2022mind} & 11.0\% & 49.0\% \\
Ozcelik et al. \cite{ozcelik2023natural} & 21.1\% & 30.3\% \\
MinD-Vis \cite{chen2023seeing} & 91.6\% & 85.9\% \\
MindEye \cite{mindeye1} & 93.6\% & 90.1\% \\
\hline
\textbf{Ours} & \textbf{95.8\%} & \textbf{95.3\%} \\
\end{tabular}
\end{table}

\noindent
\textbf{Comparison with SOTA in Brain-Image Retrieval.}  
We compare our proposed approach with the SOTA method for the Brain-Image Retrieval task. For a fair comparison, we adopt the same model architecture and training strategy as MindEye \cite{mindeye1}, with the key difference that we use the DCL loss instead of the SoftCLIPLoss used in their work.  The performance results, presented in Table \ref{tab:image_brain_retrieval}, show that our approach outperforms SOTA methods, demonstrating the effectiveness of the proposed DCL.  

\section{Conclusion}

In this paper, we have presented a bias problem in brain signals due to natural memory decay, where the bias increases with later data-collection sessions. We demonstrate, both statistically and experimentally, the impact of this issue on vision-brain understanding models. Our analysis raises concerns about how to address this problem, as previous approaches have primarily overlooked it. To tackle this problem, we have proposed a novel continual learning approach that enables the model to learn and adapt to the increasing bias across sessions. Specifically, we have introduced De-bias Contrastive Learning to mitigate bias and Angular-based Forgetting Mitigation to counteract catastrophic forgetting. Empirical results have demonstrated the effectiveness of our method in addressing bias in brain signals.

\section{Acknowledgment}

This work is partly supported by NSF CAREER (No. 2442295), NSF SCH (No. 2501021), NSF E-RISE, and NIFA Award. This research is supported by the Arkansas High Performance Computing Center, which is funded through multiple National Science Foundation grants and the Arkansas Economic Development Commission. 

\section{Data Availability}
The data that support the findings of this study are openly available at \href{https://naturalscenesdataset.org/}{https://naturalscenesdataset.org/}.

\printcredits

\bibliographystyle{cas-model2-names}

\bibliography{cas-refs}

\newpage

\vskip3pt

\bio{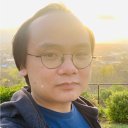}
Xuan-Bac Nguyen got PhD in Computer Science at the University of Arkansas. He received his M.Sc. degree in Computer Science from the Electrical and Computer Engineering Department at Chonnam National University, South Korea, in 2020.  He received his B.Sc. degree in Electronics and Telecommunications from the University of Engineering and Technology, VNU, in 2015. In 2016, he was a software engineer in Yokohama, Japan. His research interests include Quantum Machine Learning, Face Recognition, Facial Expression, and Medical Image Processing.
\endbio

\bio{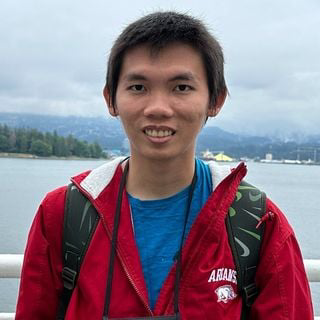}
Thanh-Dat Truong is currently a Postdoctoral Fellow at the University of Arkansas. He received his PhD degree from the University of Arkansas in 2024 and 
his B.Sc. degree in Computer Science from the University of Science (VNU) in 2019. He was a research intern in the Coordinated Lab Science program at the University of Illinois at Urbana-Champaign in 2018. 
Thanh-Dat Truong received the CVPR Doctoral Consortium Award in 2024.
His research aims to develop a robust and fair vision learning approach.
Truong's research interests include unsupervised domain adaptation, continual learning, action recognition, domain generalization, and deep generative models. His papers appear at top-tier conferences and journals such as CVPR, ICCV, NeurIPS, Neurocomputing, and ICPR.  He is also a reviewer of top-tier journals and conferences, including IEEE TPAMI, IEEE TIP, IEEE TAI, IEEE TCSVT, CVPR, ICCV, ECCV, ACCV, WACV, ICPR, ICML, ICLR, and NeurIPS.  
\endbio

\bio{pics/pawan}
Pawan Shinha is a professor of neuroscience at MIT. Pawan’s research interests span brain science, AI, and public health. His experimental work involves studying healthy individuals and also those with neurological disorders such as autism. Pawan founded Project Prakash in 2005 with the twin objectives of providing treatment to children with severe visual impairments and also understanding the mechanisms of learning and plasticity in the brain. This project has provided insights into several fundamental questions about brain function (even some that had remained open for the past three centuries) while also transforming the lives of many blind children by bringing them the gift of sight.
Pawan is a recipient of the PECASE – US Government’s highest award for young scientists, the Sloan Foundation Fellowship, and the Troland Award from the National Academies. He was inducted into the Guinness Book of World Records for creating the world’s smallest reproduction of a printed book.
\endbio

\bio{pics/khoa_luu}
is an Associate Professor and the Director of the Computer Vision and Image Understanding (CVIU) Lab in the Department of Computer Science \& Computer Engineering at the University of Arkansas, Fayetteville. He is affiliated with the UA Center for Public Health and Technology. He is an Associate Editor of the Multimedia Tools and Applications Journal, Springer Nature. He is also the Area Chair in CVPR 2023-2025, NeurIPS 2024-26, WACV 2025, ICML 2025-2026, ICLR 2025-2026, and AAAI 2026. He was the Research Project Director in the Cylab Biometrics Center at Carnegie Mellon University (CMU), USA. 
He received 8 patents, 3 Best Paper awards, and co-authored over 120 papers in conference proceedings, technical reports, and journals.
He was a vice chair of the Montreal Chapter of IEEE SMCS in Canada from September 2009 to March 2011. His research interests include Computer Vision, Semantic Video Understanding, Biometrics, Face Recognition, Tracking, Human Behavior Understanding, Domain Adaptation, Deep Generative Modeling, Image and Video Processing, Compressed Sensing, and Quantum Machine Learning. 
He is a co-organizer and chair of the CVPR Precognition Workshop 2019-2026; the MICCAI Workshop in 2019, 2020, and the ICCV Workshop in 2021.
\endbio

\end{document}